# Efficient Neural Net Approaches in Metal Casting Defect Detection

Rohit Lal[a,*], Bharath Kumar Bolla[b], Sabeesh E[c]

[a] *UpGrad Education Private Limited, Mumbai, India*
[b] *Salesforce, Hyderabad, India*
[c] *Indian Institute of Technology. Jodhpur, India*

**Abstract**

One of the most pressing challenges prevalent in the steel manufacturing industry is the identification of surface defects. Early identification of casting defects can help boost performance, including streamlining production processes. Though, deep learning models have helped bridge this gap and automate most of these processes, there is a dire need to come up with lightweight models that can be deployed easily with faster inference times. This research proposes a lightweight architecture that is efficient in terms of accuracy and inference time compared with sophisticated pre-trained CNN architectures like MobileNet, Inception, and ResNet, including vision transformers. Methodologies to minimize computational requirements such as depth-wise separable convolution and global average pooling (GAP) layer, including techniques that improve architectural efficiencies and augmentations, have been experimented. Our results indicate that a custom model of 590K parameters with depth-wise separable convolutions outperformed pretrained architectures such as Resnet and Vision transformers in terms of accuracy (81.87%) and comfortably outdid architectures such as Resnet, Inception, and Vision transformers in terms of faster inference times (12 ms). Blurpool fared outperformed other techniques, with an accuracy of 83.98%. Augmentations had a paradoxical effect on the model performance. No direct correlation between depth-wise and 3x3 convolutions on inference time, they, however, they played a direct role in improving model efficiency by enabling the networks to go deeper and by decreasing the number of trainable parameters. Our work sheds light on the fact that custom networks with efficient architectures and faster inference times can be built without the need of relying on pre-trained architectures.

*Keywords:* Efficient Deep Learning; Vision Transformers; Blurpool; Global Average Pooling; Depth-wise separable convolution; Stochastic weight averaging; Label Smoothing; Squeeze and Excite; Data augmentation; Mobilenet; Inception; Resnet; Custom models; Augmentation paradox; Mobile deployment

## 1. Introduction

Fabrication of metal sheets has been an inherent process of many steel manufacturing industries since time immemorial. A large number of modern days' appliances are manufactured using steel. Hence, maintaining a very high quality of the products and establishing a robust system to ensure quality control has become vital. Deep learning approaches have helped bridge this gap by automating defect detection and improving quality control in less time. CNN's however, have limitations in their requirement of extensive training data. Another limitation is that these architectures need massive compute power owing to their model size and the sheer number of parameters. Having huge models in production will adversely affect the outcome due to the sheer size of the model and its subsequent cascading effect on inference times. We, therefore, need to develop models that are lightweight and are also faster in inference in a production environment. Techniques for optimizing neural networks, such as depth-wise separable convolutions [1] and global average pooling [2], reduce the number of training parameters, as well as a range of other techniques [3] that increase architectural efficiency, have been studied in the past. These techniques have been shown to achieve faster inference times with reduced parameters without significant compromise in the model's accuracy with augmentations [2]. Latest architectures such as Vision transformers [3] and transfer learning architectures such as ResNet, Inception & MobileNet have been developed to handle broader datasets with reasonably good accuracies.

Click here to enter text.

However, these models have their disadvantages, the most important being their sheer size. The primary motivation of this work is to build a custom-built lightweight architecture using techniques that improve architectural efficiency and arrive at the best-performing model in inference timings without compromising on the model's performance. The research objectives are as follows

- To evaluate the role of depth-wise separable convolutions and Global Average Pooling.
- To evaluate augmentations and architectural enhancement techniques in model performance.
- To evaluate the superiority of custom-built models over pre-trained architectures.
- To build lightweight architectures that have comparable performance to pre-trained architectures with lesser inference time

The rest of this paper discusses related work followed by research methods, results and concludes with a discussion on the impact of current work and future work

## 2. Literature Review

### 2.1. Architectural Efficiency

Various research has been done in the recent past on increasing the efficiency of neural networks by improving architectural efficiency and the application of augmentations. These modifications to the neural network have helped improve the accuracy of a model, including its latency. Deep learning networks such as Mobilenet [4] and NasNetMobile [5] were designed specifically for low computational devices to enable them to be faster. Despite these networks' huge parameters, their architectural efficiency makes them suitable for deployment on edge devices. The Xception network [1] was the first to adopt depth-wise separable convolution, which was later incorporated into the Mobilenet architecture. These convolutions significantly reduced the number of parameters while enabling a network to go deeper. In order to improve the efficiency of transfer learning, layers of the transfer learning network need to be increasingly fine-tuned until the model returns the best possible accuracy for that specific dataset [6]. The number of layers to be fine-tuned varied with different architectures. In addition, recent techniques such as Blurpool and Mixup were found to improve [7] architectural efficiency by a large margin with minimal parameters. The experiments on standard datasets such as MNIST and CIFAR10 have widespread applications in real-world scenarios. Even in cases of casting defect detection, similar research has been done on evaluating the supremacy of custom-built models over pre-trained networks [8], throwing light on the fact that transfer learning may not always be ideal. The global average pooling (GAP) layer at the end of the convolutional layers was initially suggested in the "network in-network" article, which proved the GAP layer might be less prone to overfitting [2] than fully linked layers. GAP layer's ability to generalize a model with good model performance has been shown in tasks such as [9] teeth classification and the diagnosis of faults in industrial gearboxes [10].

### 2.2. Augmentations

Augmentations were used to improve a model's predictive power by generalizing its ability to predict on a wider dataset. They predominantly have a synergistic effect on a model's performance, as evident from this famous study which surveyed a wide variety of basic augmentation techniques [11]. They have also been empirically used in the detection of skin melanomas and breast MRIs [12], classification of calcaneal fractures[13], and predicting a patient's clinical condition based on a clinical history [14]. It is pertinent to note that a simple-augmentation techniques may perform equally well as specific advanced methods such as Generative Adversarial networks [15], and this nature of augmentation has been studied in this work. Efficient deep learning methods were also devised in regression scenarios in the facial key point detection [16], where manually optimized custom models were built by applying various augmentation techniques. It was shown that custom networks outperformed pre-trained neural networks, and augmentations performed superior to different imputation techniques incorporated in the experiments. Further, the inference time was found to be the least in Mobilenet V2 due to the network's architectural efficiency. However, the most promising results of combining depth wise separable convolutions with augmentations were recorded in work done by [2], where depth wise convolutions performed better than 3x3 convolutions on augmentation at higher numbers of parameters. The same work also proved that techniques such as Cutouts and Mixup fared better and consistently superior to other techniques across various architectures and parameters. Further, the paper states that a simple augmentation technique such as Random flip may perform better if not equally well as that of cutouts, considering the computational requirement that an advanced augmentation technique might incur during training. The effect of augmentations may not always be synergistic to the model performance. It is critical to apply appropriate augmentation techniques, as augmentations can sometimes distort the dataset to the point where there is a lesser distinction between the classes in the augmented dataset, resulting in underperformance of the model [8] with reduced model accuracies. – The effect of the augmentation paradox.

## 3. Methodology

Two popular frameworks, Tensorflow & Pytorch, have been used in model building and conducting various experiments [1]. In addition, the Mosaic ML library [17]  has been used to implement techniques that enhance architectural efficiency and also to perform augmentations. The experiments have been conducted on Google Colab Pro GPU instance that has 16gb vRAM.

### 3.1. Dataset

The dataset used in this research is the Kaggle GC10-DET [18]. The dataset consists of grayscale images of dimensions 28x28 with ten defect categories: punching (Pu), water spot (Ws) and weld line (Wl), rolled pit (Rp), crease (Cr), crescent gap (Cg), inclusion (In), oil spot (Os), silk spot (Ss), waist folding (Wf). There are a total of 32000 data points distributed in 10 classes, with 16000 training and 160000 test samples. The illustrations are shown in Fig 1.

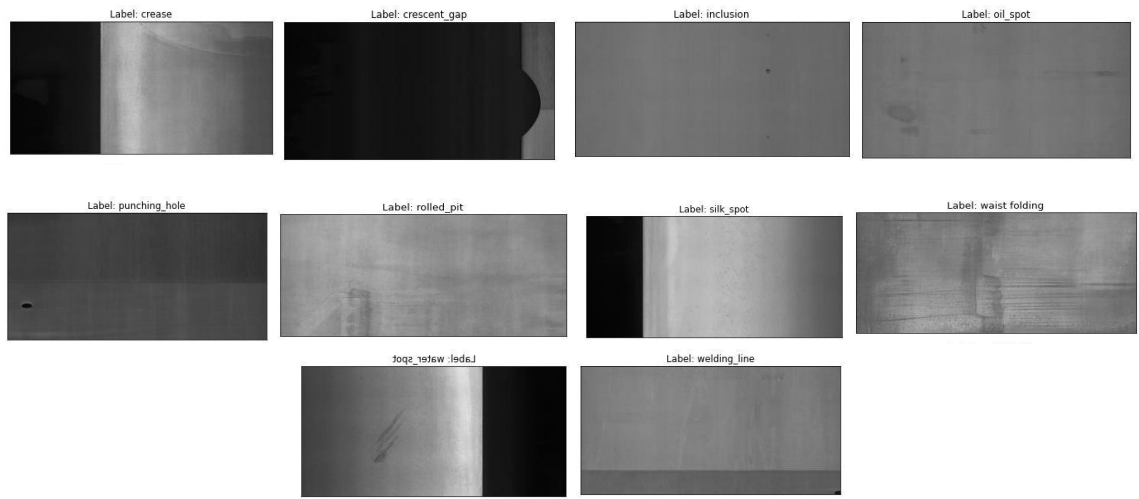

Fig 1. Ten-class distribution of the casting defects

### 3.2. Augmentations

As casting defects in practicality involve highly specific and minuscule changes in the overall metal structure, there is a possibility that the application of augmentation may result in a negative effect on the overall model performance, especially in multi-class classification. But, we still applied these augmentation techniques to prove this hypothesis.

---

### 3.2.1. Baseline augmentations

The experiments employed baseline augmentations such as random horizontal flip, random rotation, random gaussian blur, random vertical flip, shift scale rotate, random crop, random brightness, and contrast. A representation of the augmentations is shown in Fig 2.

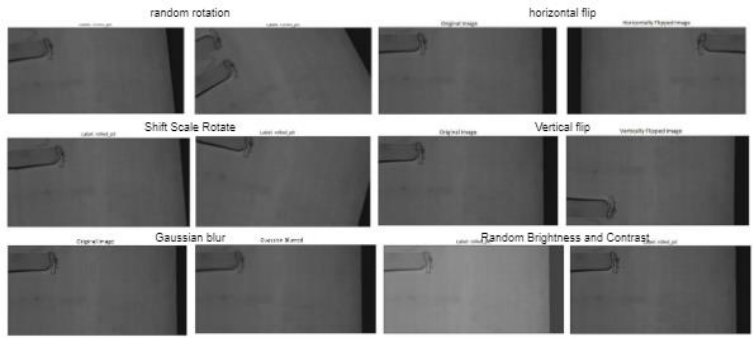

Fig 2. Baseline augmentations

### 3.2.2. Advanced augmentations – Cutouts, Mixup

A cutout (Fig 3) is a type of augmentation where random dimensions of pixels are removed from the training image and fed into the training network. This augmentation is a model generalization technique that exhibited improved model performance. Mixup is a technique where a new virtual distribution $\hat{x}$ and $\hat{y}$ is created from existing training variables $x_i$ and $y_i$. The mixup is controlled by a hyperparameter $\delta$ which determines the ratio in which the input images are mixed (1)(2). A representation of these two techniques is shown in Fig 4.

$$\hat{x} = \delta x_i + (1 - \delta) x_j \tag{1}$$
$$\hat{y} = \delta y_i + (1 - \delta) y_j \tag{2}$$

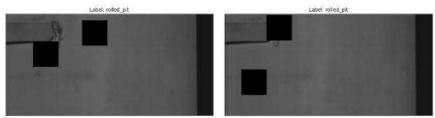

Fig 3. Cutout of dimension 5x5

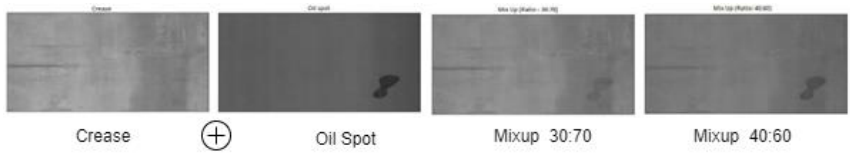

Fig 4. A mixup in varying proportions

### 3.2.3. Architectural enhancement – Label smoothing, Squeeze and excite, Blurpool

Label smoothing is a method that adds noise to the target variable, controlled by a factor called alpha. This technique is applied only in the case of cross entropy loss functions, which helps smoothen the labels, thereby acting as a regularization technique. The mathematical equation (3) for label smoothing is shown below, where K is the number of classes and is the hyper parameter.

$$y_{label\_smoothed} = (1 - \alpha) * y_{hot} + \alpha / K \tag{3}$$

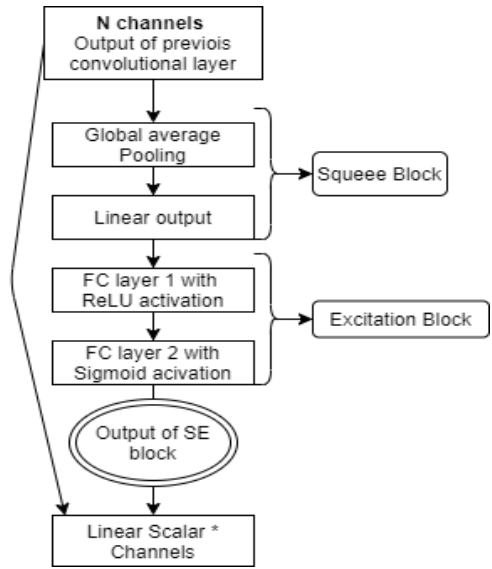

Fig 5. Squeeze and Excite Block

In squeeze and excite blocks, the output of the convolutional layer is linearized to a single-dimensional vector using global average pooling (squeeze mechanism) and then sent through two MLP layers with sigmoid activation to make a linear scalar weight (Fig 5). The weights are later applied to the individual channels of the convolutional output (attention mechanism). Blurpool (Fig 6) utilizes a kernel or a filter before the pooling operations. The kernel has an anti-aliasing effect on the input images the network is trained on, thereby resulting in increased model robustness and generalizability.

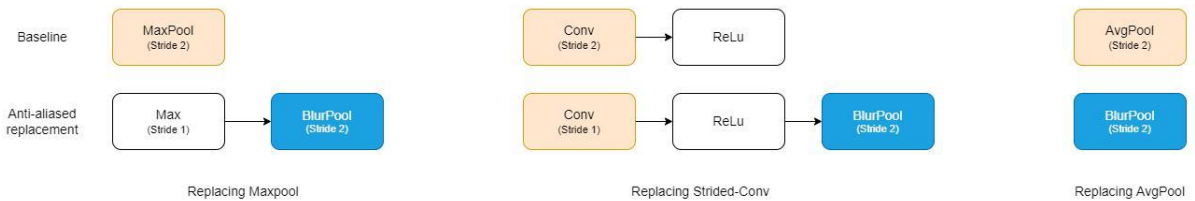

Fig 6. Blurpool

### 3.2.4. Decreasing Parameter Count – Global Average Pooling, Depth Wise Separable convolution

Depth-wise separable convolutions enable a neural network to go deeper with a decreased number of parameters in comparison with a 3x3 convolution. Depth-wise separable convolutions consist of a 3x3 convolutional function on each input channel followed by a point-wise convolution, as shown in Fig 7.

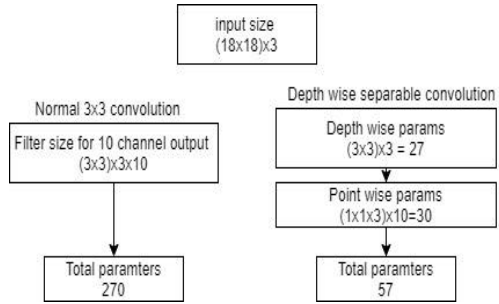

Fig 7. Depth Wise Separable Convolutions

Global average pooling (GAP) linearizes a convolutional layer's output by averaging each channel's output, resulting in a single-dimensional vector.

*3.3. Model Architecture and Evaluation*

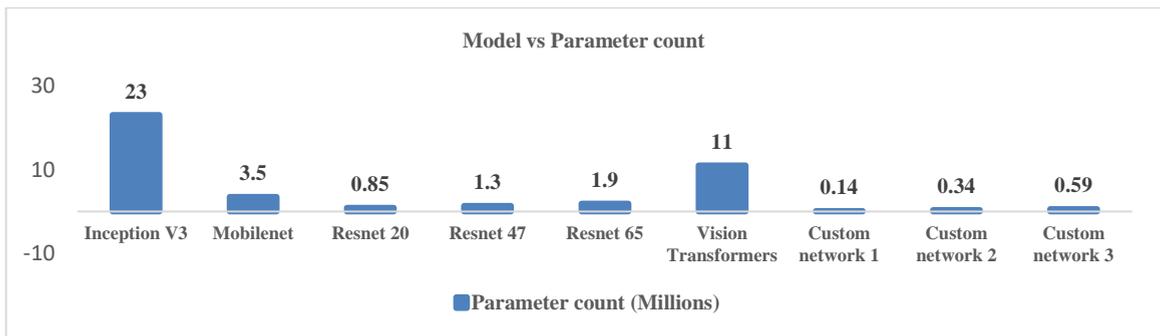

Fig 8. Model vs. Parameters

Three baseline pretrained and bespoke models have been employed in the given classification task. The 3x3 and depth-wise separable convolutions were used to reduce the number of parameters in the custom models, which were then developed sequentially. We used accuracy and inference time as metrics to measure the model's performance. The models are summarized in Fig 8.

*3.4. Loss function*

Given that there are multiple classes involved in this scenario, the loss function (4) utilized here is a cross entropy loss function.

$$Cross\ Entropy\ Loss = -\sum_{i=1}^{k} y_i \log(\hat{y}_i) \quad (4)$$

# 4. Results

The superiority of bespoke models over pre-trained networks in terms of accuracy and inference times is discussed in depth further below.

### 4.1. Depth-wise vs 3x3 convolutions

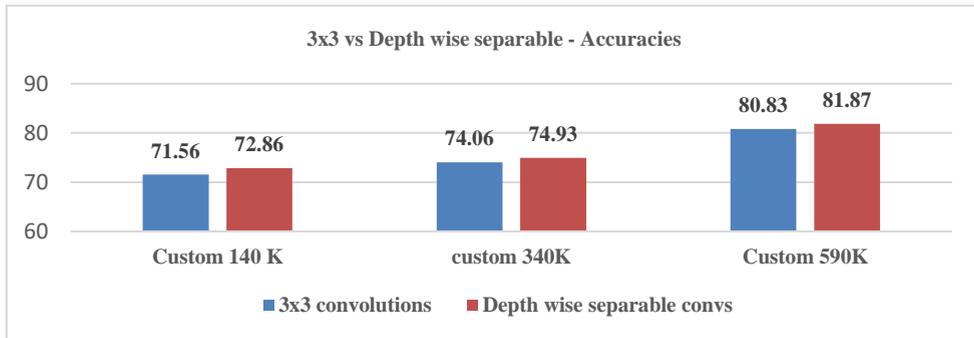

Fig 9. 3x3 vs. Depth Wise Separable Convolutions - Accuracies

Architectural saturation is observed when there is no substantial improvement in accuracy after optimizing all convolutional layers in terms of the number of channels, filters, and depth of a network using a given set of functions. In the current work, we observed that architectures with **depth-wise separable convolutions perform better than architectures with 3x3 convolutions**. This phenomenon is evident in Fig 9, with an increase in accuracy of 1.3%, 0.87%, and 1.04% for the 140K, 390K, and 590K architectures. The better performance of these models can be attributed to the ability to go deeper in contrast with their 3x3 counterparts, as seen in Table 1.

### 4.2. Architectural Efficiency

The best performing custom model with depth-wise separable convolutions was subjected to a series of experiments (Table 2) involving optimization by techniques such as Blurpool, Squeeze & Excite, Mixup, and Stochastic weight averaging to study the effect of network performance. It was observed that improving architectural efficiency using these techniques improved the model performance. **Blurpool outperformed other methods** with an accuracy increase of 2.11% (83.98%) from the baseline model. The accuracy of the model is on par with SOTA architectures such as Mobilenet (86.38%) and Vision Transformers (85.74%) but with a significant **reduction in the number of training parameters** (**5.93 times & 18.64 times,** respectively)

Table 1. Architectural Optimizations

| Model techniques | Accuracy |
| --- | --- |
| **Custom model (590 k params) – 3x3** | 80.83 |
| **Custom model (590 k params) – Depth wise separable convolutions** | 81.87 |
| Blurpool (BP) | **83.98%** |
| **Stochastic Weight averaging (SWA)** | 83.18% |
| **Squeeze and Excite (SE)** | 82.98% |
| **Mixup** | 80.98% |
| **BP+SE+SWA** | 83.99% |

### 4.3. The superiority of Custom Models over Pre-trained networks

The accuracy of the various models built has been summarized in Table 1. From Table 1 and Table 2, it is observed that Inception, MobileNet V2, and Vision Transformers have higher accuracies, but they have been achieved at the cost of a higher number of parameters. In contrast, custom models with only 590 k parameters (46 times lesser) perform better than pre-trained architectures such as Resnet architectures with an accuracy of 81.87%.

Table 2. Model Accuracies

| Model | Architecture | Number of conv layers | Test Accuracy |
|---|---|---|---|
| Inception | Pre-defined | Pre-defined | 91.48% |
| Mobilenet | Pre-defined | Pre-defined | 86.38% |
| Vision Transformers | Pre-defined | Pre-defined | 85.74% |
| Custom (0.59 M params)- 3x3 | 8(Conv3) + GAP + Classification | 8 | 80.83% |
| Custom (0.59 M params) – Depth wise | 2(Conv3) +1(ConvDW) + 1(Conv3) + 1(ConvDW) + 1(Conv3) + 6(ConvDW) + GAP + Classification | 12 | 81.87% |
| Resnet-20 | Pre-defined | Pre-defined | 70.63% |
| Resnet-47 | Pre-defined | Pre-defined | 71.06% |
| ResNet-65 | Pre-defined | Pre-defined | 74.04% |
| Custom (0.34 M params) - 3x3 | 7(Conv3) + GAP + Classification | 7 | 74.06% |
| Custom (0.34 M params) - Depth Wise | 2(Conv3) +1(ConvDW) + 1(Conv3) + 1(ConvDW) + 1(Conv3) + 5(ConvDW) + GAP+Classification | 11 | 74.93% |
| Custom (0.14 M params) - 3x3 | 7(Conv3) + GAP + Classification | 7 | 71.56% |
| Custom (0.14 M params) - Depth wise | 2(Conv3) +1(ConvDW) + 1(Conv3) + 1(ConvDW) + 1(Conv3) + 3(ConvDW) + GAP+Classification | 9 | 72.86% |

*\* Conv3 – Convolutional layer with 3x3 kernel / ConvDW – Conv layer with depth-wise-separable convolutions*

### 4.4. Augmentation paradox

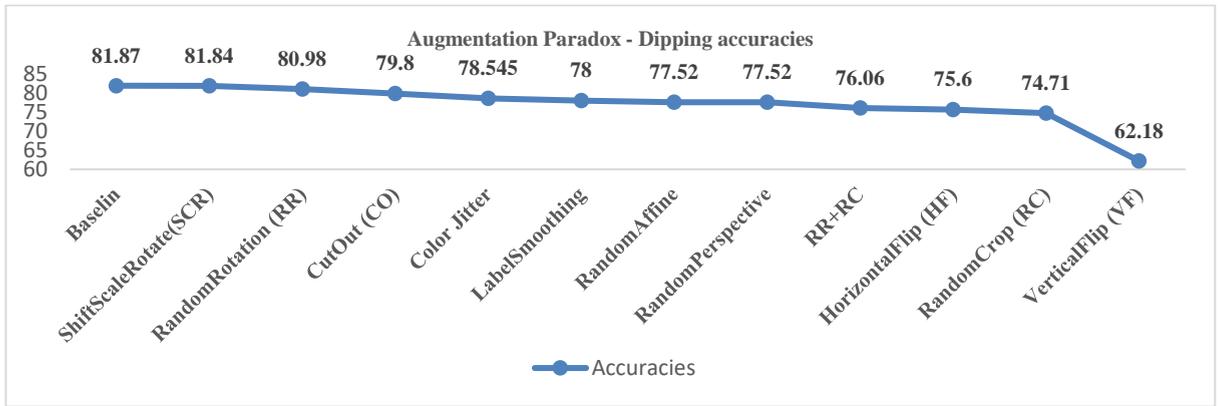

Fig 10. Augmentation Paradox - Negative effects

Various Augmentation techniques were applied to the custom model with 590 K parameters and depth wise convolutions, as shown in Fig 10. We noticed that all augmentations have a paradoxical effect on the model's performance resulting in reduced accuracies. This behavior might be hypothesized as an increased number of classes resulting in distortion of the original distribution and reduced class distinction post augmentation.

Table 3. Inference time analysis

| Models | Parameters | Parameter ratio | Inference time in ms (CPU) | | |
|---|---|---|---|---|---|
| | | | Batch latency | Individual latency | Individual latency |
| **Inception** | 23 M | 1 x | 79 | 54 | 1.00 x |
| **Vision Transformer** | 11 M | 2 x | 52 | 28 | 1.93 x |
| **MobileNet** | 3.5 M | 7 x | 47 | 25 | 2.16 x |
| **Resnet-20** | 0.85 M | 27 x | 35 | 13 | 4.15 x |
| **Resnet-47** | 1.3 M | 18 x | 40 | 14 | 3.86 x |
| **ResNet65** | 1.9 M | 12 x | 45 | 15 | 3.60 x |
| Custom 3 - 3x3 convs | **590 K** | **46 x** | **33** | **12** | **4.50 x** |
| Custom 3 - depth-wise convs | **590 K** | **46 x** | **33** | **12** | **4.50 x** |
| **Custom 2 - 3x3 convs** | 340 K | 68 x | 31 | 11 | 4.91 x |
| **Custom 2 - depth-wise convs** | 340 K | 68 x | 31 | 11 | 4.91 x |
| **Custom 1 - 3x3 convs** | 140 K | 164 x | 29 | 9 | 6.00 x |
| **Custom 1 - depth-wise convs** | 140 K | 164 x | 29 | 10 | 5.40 x |

We measured inference time on the CPU machine, and both batch-wise latency and individual latency were recorded (Table 3). While the convolution type does not seem to have any significant effect on the latency of the model, it is observed that they are directly proportional to the number of parameters, as more parameters require more computational time during the forward pass of the model. Hence, the custom model with 590 K parameters has less latency than any pre-trained models. Table 3 custom models perform four to five times faster than the pre-trained architectures such as Inception, including vision transformers.

## 5. Conclusion

The work pivots around the study of custom-built lightweight neural networks and pre-trained neural networks such as Resnet, Mobilenet, and Inception networks to evaluate the supremacy of the custom models. We used efficient deep learning techniques such as depth-wise separable convolutions, global average pooling layer, and advanced architecture optimizations such as Blurpool, Stochastic Weight averaging, Mixup, Label smoothing, and Squeeze and Excitation on custom-built models. This work is one of the first studies to employ all these techniques on a custom model. Our custom models performed superior (Accuracy of 81.87 %) to pre-trained networks with the minimal number of parameters (46 times lesser), hence suggesting that pre-trained models may not always be ideal in a given machine learning task. Furthermore, architectures with depth-wise separable convolutions had a definitive edge over architectures with 3x3 convolutions in model accuracies due to their ability to generate a deeper architecture. On average, architectures with depth-wise separable convolutions had a 1.07% increase in accuracy.

While these architectures significantly reduce the number of training parameters, they did not cause any significant increase in the model compared with 3x3 convolutions. The latency of a model was found to be directly proportional to the number of parameters. Custom models performed four to six times faster than Inception and Vision transformers and two times faster than Mobilenet. Blurpool was the best performing architecture optimizing algorithm, improving model accuracy by 2.11% and yielded similar results as SOTA architectures such as Mobilenet and Vision transformers. The effect of augmentations was detrimental to the model performance due to many classes and similarities among the defects, resulting in distortion of the original distribution. Hence, appropriate augmentation must be chosen for a given scenario after analyzing the dataset. The outcome of this research further re-establishes the findings of our earlier work performed on standard datasets such as MNIST and CIFAR-10 [2] [7].[2] Similar results were observed on casting defect identification [8], wherein the authors were able to build lightweight models (100 times lesser number of parameters) with improved inference time (four to six times faster in comparison with MobileNet and NasNetMobile). The findings of this research prove beyond doubt a stronger testament to the earlier work done, as we were able to achieve convincing evidence of the same in a multi-class classification scenario in contrast to the binary-class classification scenario of the earlier works. The future scope of work includes implementing other techniques such as quantization, pruning and knowledge distillation.